\documentclass{vgtc}                          % final (conference style)
\ifpdf%                                % if we use pdflatex
  \pdfoutput=1\relax                   % create PDFs from pdfLaTeX
  \usepackage{graphicx}                % allow us to embed graphics files
  \DeclareGraphicsExtensions{.pdf,.png,.jpg,.jpeg} % for pdflatex we expect .pdf, .png, or .jpg files
\else%                                 % else we use pure latex
  \usepackage{graphicx}                % allow us to embed graphics files
  \DeclareGraphicsExtensions{.eps}     % for pure latex we expect eps files
\fi%

\graphicspath{{figures/}{pictures/}{images/}{./}} % where to search for the images

\usepackage{microtype}                 % use micro-typography (slightly more compact, better to read)
\PassOptionsToPackage{warn}{textcomp}  % to address font issues with \textrightarrow
\usepackage{textcomp}                  % use better special symbols
\usepackage{mathptmx}                  % use matching math font
\usepackage{times}                     % we use Times as the main font
\usepackage{cite}                      % needed to automatically sort the references
\usepackage{tabu}                      % only used for the table example
\usepackage{booktabs}                  % only used for the table example
\usepackage{mwe}                       % used to generate placeholder figures
\usepackage{dblfloatfix}
\usepackage{float}
\usepackage{amsmath}
\usepackage{enumitem}

\onlineid{8234}

\vgtccategory{Research}

\vgtcinsertpkg

\title{Enhancing Perception and Immersion in Pre-Captured Environments through Learning-Based Eye Height Adaptation}

\author{Qi Feng\thanks{e-mail: fengqi@ruri.waseda.jp}\\ %
        \scriptsize Waseda University %
\and Hubert P. H. Shum\thanks{e-mail: hubert.shum@durham.ac.uk}\\ %
     \scriptsize Durham University %
\and Shigeo Morishima\thanks{e-mail: shigeo@waseda.jp\vspace{-4em}}\\ %
     \parbox{1.4in}{\scriptsize \centering Waseda Research Institute \\ for Science and Engineering}}

\teaser{
  \centering
  \vspace{-1.3em}
  \includegraphics[width=\linewidth]{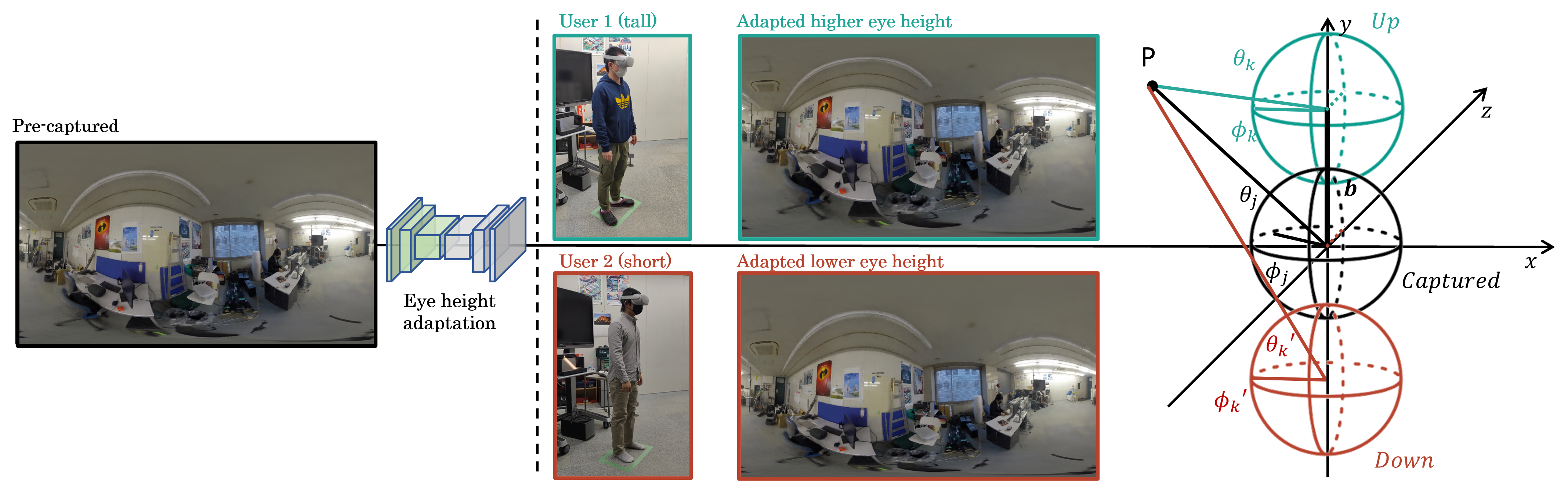}
  \vspace{-2em}
  \caption{%
  	We propose a learning-based approach for generating novel views from pre-captured omnidirectional inputs, which can adapt to the user's eye height during playback, resulting in an improved perceptive and immersive experience.  %
  }
  \label{fig:teaser}
}

\abstract{Pre-captured immersive environments using omnidirectional cameras provide a wide range of virtual reality applications. Previous research has shown that manipulating the eye height in egocentric virtual environments can significantly affect distance perception and immersion. However, the influence of eye height in pre-captured real environments has received less attention due to the difficulty of altering the perspective after finishing the capture process. To explore this influence, we first propose a pilot study that captures real environments with multiple eye heights and asks participants to judge the egocentric distances and immersion. If a significant influence is confirmed, an effective image-based approach to adapt pre-captured real-world environments to the user's eye height would be desirable. Motivated by the study, we propose a learning-based approach for synthesizing novel views for omnidirectional images with altered eye heights. This approach employs a multitask architecture that learns depth and semantic segmentation in two formats, and generates high-quality depth and semantic segmentation to facilitate the inpainting stage. With the improved omnidirectional-aware layered depth image, our approach synthesizes natural and realistic visuals for eye height adaptation. Quantitative and qualitative evaluation shows favorable results against state-of-the-art methods, and an extensive user study verifies improved perception and immersion for pre-captured real-world environments.%
} % end of abstract

\CCScatlist{
  \CCScatTwelve{Computing methodologies}{Computer graphics}{Graphics systems and interfaces}{Mixed / augmented reality};
  \CCScatTwelve{Computing methodologies}{Computer graphics}{Image manipulation}{Image-based rendering}
}

\begin{document}



\maketitle
\section{Introduction} 
Pre-captured immersive content for virtual reality (VR) has gained increasing attention from the commercial and research communities for its potential applications in fields such as medicine, education, entertainment, and prototyping \cite{shimamura2020audio}. Omnidirectional cameras capture egocentric perspectives that provide greater immersion than traditional media, fostering more effective interactions between the content and the user \cite{feldstein2020egocentric}. By directly recording the environment and simulating real-world perceptions during playback, pre-captured content offers more photo-realistic cognitive stimuli than model-based virtual environments. However, this also makes post-processing of the visuals more challenging compared to traditional model-based virtual environments.

Previous studies have shown that manipulating the visual eye height within virtual environments can significantly affect distance perception \cite{leyrer2011influence}, and that virtual and real environments can elicit similar visual responses \cite{feldstein2020egocentric}. However, the influence of eye height on perception in pre-captured immersive content has not been studied, due to the difficulty of freely altering the eye height after the footage is captured. If a significant influence on perception and immersion is found, an effective approach to adapt the eye height of the user for existing pre-captured immersive content would be highly desirable.

Traditional image-based reconstruction methods often require specific capture setups with a sufficient number of inputs for baselines \cite{flynn2016deepstereo,hedman2017casual}. %,hedman2018deep}. 
Novel view synthesis typically uses multi-layered image representations combined with depth-based warping algorithms \cite{srinivasan2019pushing,zhou2018stereo} to replicate parallax effects, resulting in holes in occluded regions. To address missing information, recent convolutional neural network (CNN)-based approaches use light field data \cite{mildenhall2019local,kalantari2016learning}, piece-wise planar images \cite{huang2014image}, or local inpainting \cite{iizuka2017globally}. These methods have shown great potential for manipulating the eye height in our application, but most of them are designed for pinhole cameras and do not perform well with 360-degree inputs due to irregular distortions introduced by equirectangular projection \cite{zioulis2018omnidepth, feng2022360}. % \cite{han2020piinet}.

In this paper, we first propose a pilot study to verify whether different eye heights have a significant influence on users' perception and immersion when viewing pre-captured real-world environments, a hypothesis that has not previously been tested. Unlike virtual environments where the eye height can be easily adjusted, we capture identical scenes at multiple eye heights under controlled conditions to optically simulate different eye levels in the real world using state-of-the-art equipment. The results of the %pilot 
study show an improved perception and immersion, providing the basis for the subsequent eye height adaptation system. It %This study 
contributes to both future application designs and a better understanding of human perceptions.

Motivated by the pilot study, we propose a learning-based approach for adapting the eye height of pre-captured immersive content. The system consists of a depth estimation stage and an inpainting stage. We first introduce a novel omnidirectional-aware multitask architecture that learns depth and semantic segmentation in two formats, enabling the network to generate high-quality depth and semantic segmentation that facilitates the inpainting stage for 360-degree input. In the inpainting stage, we improve upon the existing layered depth image (LDI) approach \cite{shih20203d} by using the omnidirectional-aware depth and semantic segmentation information to guide the synthesis of natural and realistic textures for occluded regions, enabling eye height adaptation for pre-captured real-world environments.

With quantitative and qualitative evaluation, our experimental results shows an improved performance over existing methods, showing that the proposed method is able to generate eye height-adapted results with satisfying quality and efficiency. An extensive user study further verifies the effectiveness of our learning-based approach in improving user perception and immersion for pre-captured immersive content in VR. We believe that its application can benefit a wide range of existing pre-captured media in 360-degree format for better immersion and experience.

To summarize, our contributions are as follows:
\begin{enumerate}[noitemsep,topsep=0pt]
\item We propose the first pilot study to show a significant influence of altered eye heights on perception and immersion for pre-captured immersive content with real environments;
\item We propose a two-stage approach for omnidirectional-aware eye height adaptation. A novel network estimates accurate depth and semantic segmentation, and the following inpainting stage improves the layered depth image approach with guides to synthesize high-quality visuals;
\item The implementation and the main user study validate the effectiveness of image-based eye height adaptation in improving users’ perception and immersion in pre-captured real environments within VR.
\end{enumerate}

\section{Related Work}

\subsection{Perception and Immersion in Virtual Reality}

\textbf{Perception and immersion.} In recent years, a body of VR research has sought to identify the factors that cause the distance compression effect often observed in VR. This helps resolve egocentric perceptual deficiencies and improve future VR designs. These studies typically use verbal estimates \cite{witmer1998judging,feldstein2020egocentric,leyrer2011influence}, blind walking \cite{kelly2017perceived,jones2016vertical}, and perceptual matching tasks \cite{li2011underestimation} to assess distance perception. Verbal estimates require participants to report the absolute distance of a target numerically, while blind walking also involves participants' perceptual-motor skills \cite{maruhn2019measuring,schneider2018locomotion}. Perceptual matching tasks typically investigate the ordinal depth of multiple objects. Although different tasks have their own advantages for investigating perception in VR, verbal estimates are reported to remain consistent across a wide range of distances \cite{klein2009measurement,loomis2008measuring} and environments, regardless of whether they are modeled virtual environments with lower visual fidelity or photo-realistic captured scenes \cite{feldstein2020egocentric}, which is crucial for this work.

Egocentric perception in VR is typically influenced by human, technical, and environmental factors. In terms of human factors, %previous research has shown that 
physical characteristics such as gender \cite{creem2005influence}, age \cite{murgia2009estimation}, and height \cite{murgia2009estimation} do not significantly affect distance perception, and prior experience with VR does not improve distance estimation accuracy \cite{murgia2009estimation}. However, the feeling of presence, or immersion in VR, has been found to influence judgments \cite{jones2012improvements,jones2011peripheral}. One possible reason is the poor fit of the head-mounted display (HMD) during experiments. To address this issue, we include a presence survey in our user studies to assess perception in VR and evaluate the experienced immersion for eye height adaptation.

Technical factors also play a role in influencing egocentric perception in VR. For instance, newer hardware systems that provide a larger Field of Views (FOVs) have been shown to improve accuracy in a range of tasks \cite{buck2018comparison,kelly2017perceived,young2014comparison}. In addition, ergonomic design can alleviate the distance compression effect \cite{buck2018comparison,jones2008effects,willemsen2004effects,willemsen2009effects}. High display resolution also contributes to better visibility and improved presence and perception \cite{feldstein2020egocentric}. Stereoscopic vision provides depth cues through disparity \cite{cutting1995perceiving}, but its effectiveness diminishes for distant targets \cite{lappin2006environmental,palmisano2010stereoscopic}. For tasks beyond proximity, stereoscopic vision does not offer a clear advantage over monocular vision \cite{willemsen2008effects,creem2005influence}. 
% In this research, we %choose to use utilize state-of-the-art capture devices to provide stereoscopic environments, as most existing pre-captured immersive content is already stereoscopic, which makes it easier to apply the proposed eye height adaptation to a wider range of immersive media.

Environmental factors, such as the realism and composition of the virtual environments, also play a role in influencing egocentric perception in VR \cite{murgia2009estimation,surdick1997perception,vaziri2017impact}. %Research has shown that 
Highly realistic environments improve accuracy compared to non-photorealistic renderings \cite{vaziri2017impact}. In non-photorealistic environments, participants consistently underestimate distances \cite{peer2017evaluating,renner2013perception,schneider2018locomotion}, whereas in real-world ones, %environments, 
estimations are often accurate %for targets up to 10m away 
\cite{renner2013perception}. We therefore %put forward the 
hypothesize that adapting the eye height for captured real-world scenes would benefit perception and immersion in VR. Further research also indicates that compositional visual cues, such as linear perspective and ground textures, affect performance. %Andre and Rogers \cite{andre2006using} have shown that
Considering indoor and outdoor scenes play an important role in perception \cite{andre2006using}, % Therefore, 
we prepared %multiple environments for both indoor and outdoor 
both types of 
scenes to counteract compositional factors and study outdoor conditions, which are under-researched. %not well-researched in the literature.

\textbf{Influence of eye heights.} The eye height is a crucial source of information for egocentric distance and depth perception \cite{leyrer2011influence} \cite{feldstein2020egocentric} \cite{masnadi2022effects}. By observing the proportion of the horizon occluded by an object, individuals can infer the height and distance of the target based on either an explicit or implicit horizon in the environment \cite{rothe2019camera}. Knowing their eye level allows individuals to estimate the distance of an object based on its size. Previous research has shown that manipulations of eye height have a significant impact on perceived distance in virtual environments \cite{leyrer2011influence}. Increasing the virtual eye height by 50 cm increases the distance compression effect \cite{ries2008effect}, while decreasing the virtual eye height does not have a significant influence on perception \cite{kelly2022distance} \cite{leyrer2011influence}. Investigating the perception of the real world is more challenging, as it is not easy to manipulate eye height for pre-captured environments \cite{feldstein2020egocentric}. Previous research mainly focus on resolution \cite{el2019survey}, FOV \cite{masnadi2021field}, realism \cite{feldstein2020egocentric}, and other HMD-related factors \cite{el2019distance} when investigating perception of pre-captured environments. One study attempts manipulating height targets with a constant eye height in real world, observing a compression of distance perception similar to virtual environment studies \cite{ooi2001distance}. Recent research also revealed that indoor/outdoor conditions affect distance perception in real environments \cite{dukes2022visual}. In this research, we use state-of-the-art capture equipment to capture real-world environments at different eye heights and a HMD to investigate the influence of eye heights on perception.

\subsection{Novel View Synthesis for Virtual Reality}

\textbf{Omnidirectional-aware novel view synthesis.} Novel view synthesis from pre-captured images is a persistent challenge in computer vision and computer graphics. Traditionally, structure-from-motion (SfM) and multi-view stereo are applied to a collection of images to estimate point clouds and camera extrinsic through geometric models \cite{penner2017soft}. However, it usually required sufficient baselines for multiple viewpoints \cite{lin2020deep}, and the computation is quite expensive to generate and represent the entire scene with detailed meshes. To address this challenge, researchers have proposed several representation methods that allow for the generation of novel views without the need for a complete 3D model. These methods include multi-plane images (MPI) \cite{mildenhall2019local, srinivasan2019pushing}, layered depth images \cite{shih20203d, li2021mine}, and light fields \cite{levoy1996light, kalantari2016learning}. However, each of these methods has its own limitations. For example, MPI representation is lightweight and can capture specular surfaces, but its discretized representation can lead to suboptimal performance for sloped surfaces \cite{waidhofer2022panosynthvr}. In addition, MPI representation with predetermined layer structures can suffer from abrupt layer changes across discontinuities in depth, leading to inferior preserved locality. LDI representation, on the other hand, allows for arbitrary depth complexity with great efficiency thanks to its sparsity. Recent work has proposed storing connectivity information between layers in LDI representation \cite{li2021mine, hedman2017casual, hedman2018instant}, which allows for the breakdown of the global inpainting problem into sub-areas that can be solved iteratively. This representation is also well-suited for omnidirectional images due to its extremely large field of view.

\begin{figure}[!h]
\centering
  \includegraphics[width=0.48\textwidth,keepaspectratio]{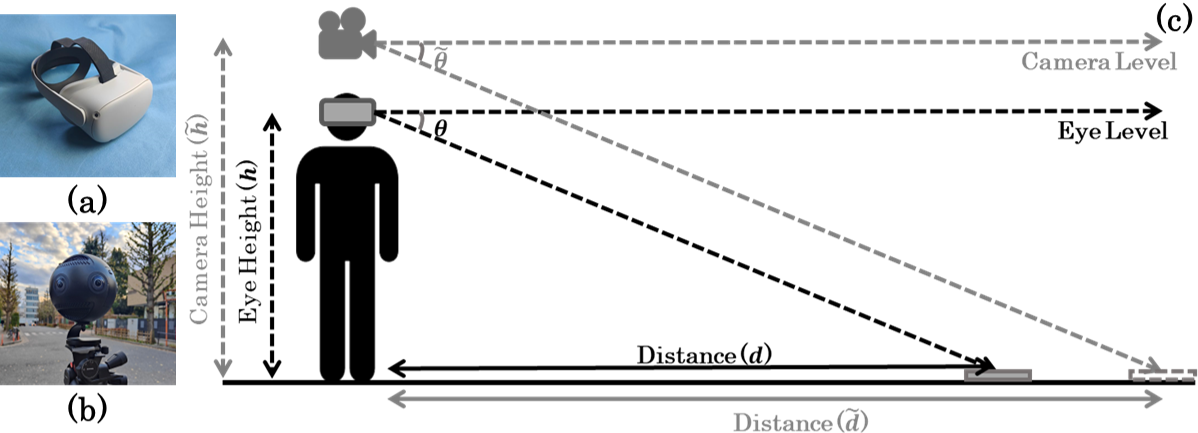}
  \vspace{-1em}
  \caption{The configuration of the pilot study: (a) HMD (Meta Quest 2). (b) Omnidirectional camera (Insta360 Pro 2). (c) Study design.}
  \label{fig:setup}
\end{figure}

\vspace{-12pt} 

\begin{figure}[htb]
\centering
  \includegraphics[width=0.45\textwidth,keepaspectratio]{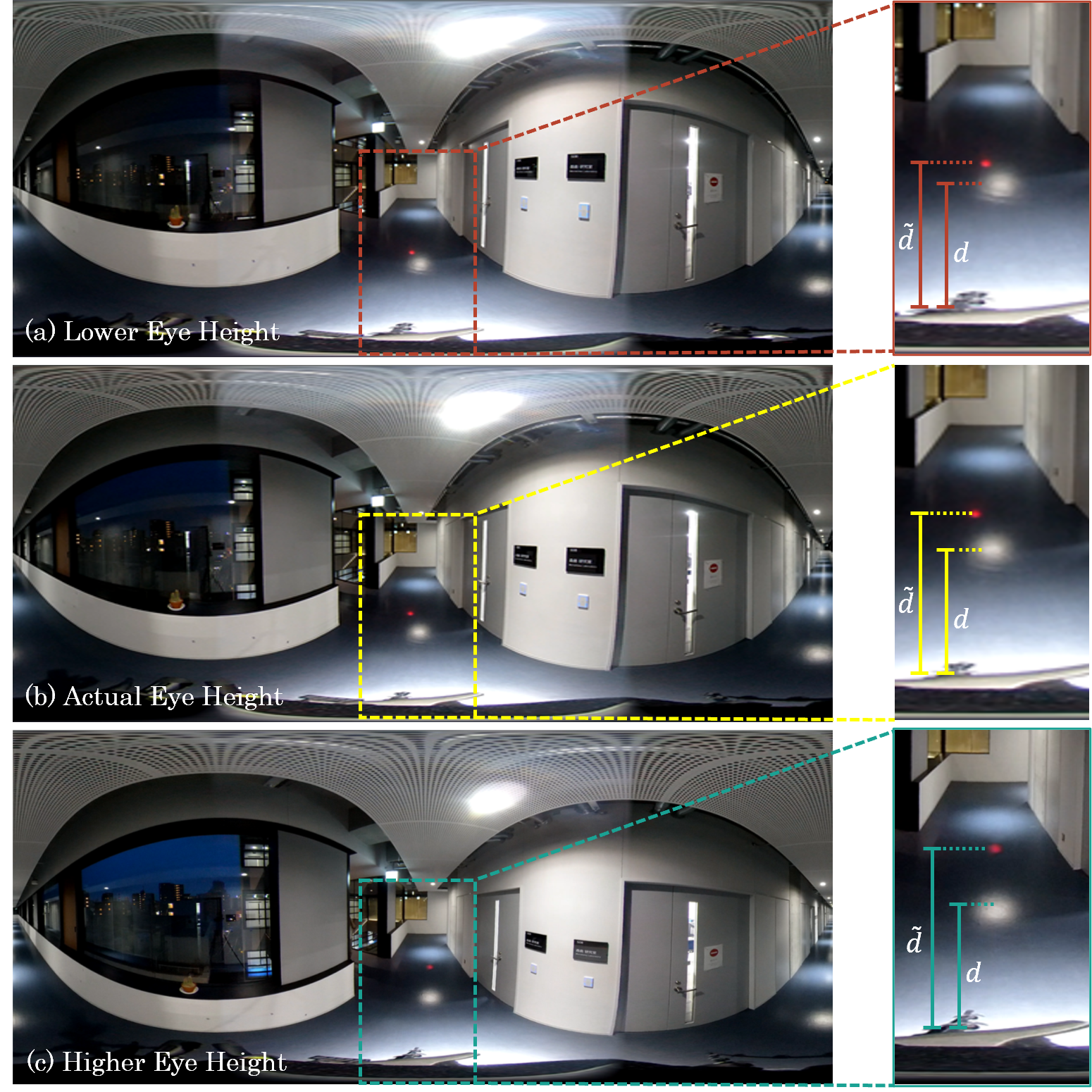}
  \vspace{-1em}
  \caption{An example of a pre-captured immersive environment with different eye heights. The distances to the white spot ($d$) and the angle of declination to the red spot ($\theta$) are kept constant across each condition. As demonstrated, the different eye heights result in discrepancies in the perception of $\tilde{d}$ in virtual environments.}
  \label{fig:environment}
\end{figure}

Recently, CNN-based approaches have been used to generate views from sparse light field data \cite{kalantari2016learning}, piecewise planar images \cite{liu2018planenet}, and neural radiance field (NeRF) \cite{mildenhall2021nerf}. Light-field photography allows for photorealistic rendering of novel views, but generally requires dense capture of the scene to achieve good results. Piecewise planar image-based approaches enable view synthesis from monocular image input, but are restricted to specific scenarios \cite{liu2018planenet}. Pathdreamer \cite{koh2021pathdreamer} uses 2D image-to-image translation to synthesize novel views for omnidirectional RGB-D images, but it focuses on a specific domain (indoor scenarios) with limited 3D consistency. NeRF \cite{mildenhall2021nerf} shows great potential with high-quality and photorealistic views for complex scenes, but it requires large amounts of consistent input images with known relative positions. Later research has proposed ways to alleviate the requirement of input \cite{xu2022sinnerf} and to reduce computational cost \cite{hong2022headnerf} \cite{deng2022fov}. For omnidirectional input, OmniNeRF \cite{hsu2021moving} learns from omnidirectional RGB-D images and shows good performance for our application. However, it relies on neighborhood interpolation to complete occluded regions, which can result in visual artifacts. Combining with additional weakness of fine details and lengthy per-frame training, we find image-based methods to be a more practical choice for this application. 

% While other 360 methods exist \cite{waidhofer2022panosynthvr} \cite{lin2020deep}, due to their input limitations, 

% but the quality of the result still highly depends on a larger number of input images and the computational cost is not suitable for real-time virtual reality applications. Furthermore,

\textbf{Image inpainting.} Image inpainting is the process of filling in missing regions of an image with plausible content. Traditional methods for inpainting include example-based methods, which transfer the texture of other regions to the missing pixels using non-parametric patch synthesis \cite{darabi2012image, huang2014image} or Markov Random Fields to propagate from the boundaries \cite{komodakis2007image}. More recent approaches have employed convolutional neural networks (CNNs) to predict semantically meaningful results by learning from large training datasets \cite{iizuka2017globally}. Further improvements to network architectures have been proposed to handle irregularly-shaped holes \cite{liu2018image} with diffusion models \cite{lugmayr2022repaint}. Two-stage approaches have also been developed that predict the structure of missing areas before completing them contextually \cite{lugmayr2022repaint}. For omnidirectional images, cubemap projection is used to represent the spherical nature of the inputs \cite{feng2022360}. Data-driven image completion for 360-degree images has shown promising results when combined with OmniNeRF \cite{hsu2021moving}, but is limited to indoor scenarios due to limited training data. In this paper, we extend the LDI-based approach with cubemap representation and adopt the two-stage approach \cite{shih20203d} for inpainting missing regions. By breaking down the large field of view into non-distorted local regions, we can solve the inpainting problem using a standard CNN.

\begin{figure}[htb]
\centering
  \includegraphics[width=0.45\textwidth,keepaspectratio]{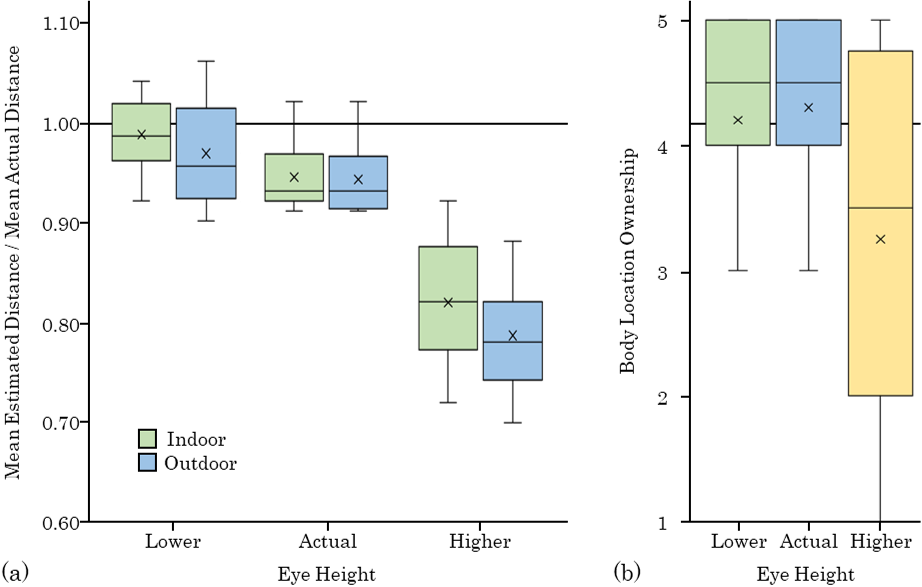}
  \vspace{-1em}
  \caption{Pilot study results in distance perception and immersion.}
  \label{fig:pilot}
\end{figure}

\vspace{-8pt} 

\begin{figure*}[b]
\centering
  \vspace{-1em}
  \includegraphics[width=0.88\textwidth,keepaspectratio]{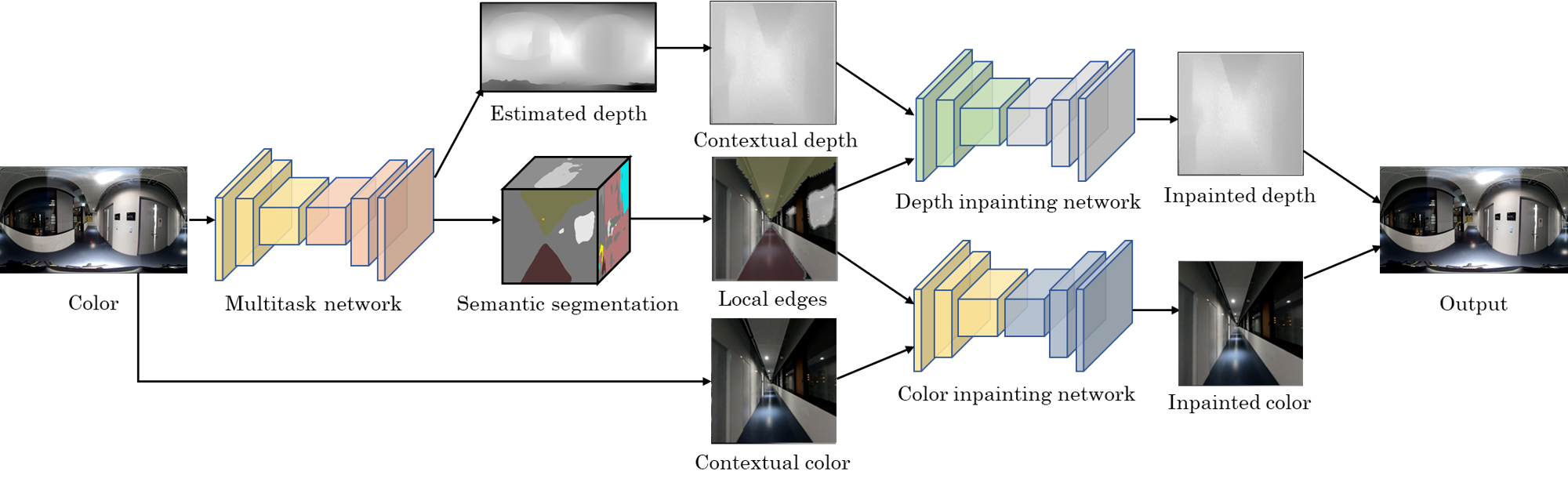}
    \vspace{-1em}
  \caption{The overview of the proposed eye height adaptation approach. The proposed eye height adaptation approach is %involves 
  a multitask architecture that leverages both equirectangular and cubemap projections to predict depth and semantic segmentation. The architecture is omnidirectional-aware and uses semantic segmentation as guiding information to complete the layered depth and color images using inpainting networks. The final step involves synthesizing natural and realistic visuals that are adapted to different eye heights for improved perception and immersion.}
  \label{fig:overview}
\end{figure*}

\section{Pilot Study: Eye Height in Pre-Captured Real Environments}

\textbf{Study concept.} The aim of this pilot study is to examine the potential influence of eye height on perception and immersion in pre-captured immersive environments. It has been suggested that the more closer the visuals in a virtual environment match real-world experience, the greater the sense of presence in that environment \cite{feldstein2020egocentric}. While the influence of altered eye height in model-based virtual environment has been verified \cite{leyrer2011influence}, it has yet to be confirmed for pre-captured real environments. 

\textbf{Design and tools.} 
%In this study, 
We conducted experiments that consisted of two stages: capture and playback. During the capture stage, we used a state-of-the-art omnidirectional camera to record real-world environments with high visual fidelity. In the playback stage, participants used a HMD to view the pre-captured environments in an immersive way. The HMD used in the experiment was the Oculus Quest 2, which had a display resolution of 1832 $\times$ 1920 per eye and a refresh rate of 90 Hz. The effective horizontal and vertical FOV of the HMD was 90$^{\circ}$ $\pm$ 5$^{\circ}$ and 93$^{\circ}$ $\pm$ 5$^{\circ}$, respectively. This difference was due to the varying distances between participants' eyes and the lens of the HMD (e.g., whether they were wearing glasses or not). The interpupillary distance used during the experiments was $6.5 cm$, as suggested by previous studies \cite{thompson2004does}. During playback, we used a basic omnidirectional video playback function in the HMD, which allowed for 3-degrees-of-freedom interactions between the participants and the environment.

To assess perception, participants were asked to provide verbal estimates of the distance of a truncated cone-shaped target with a height of $10 cm$ in multiple pre-captured environments with varying eye heights. The dimensions of the target were not disclosed to participants in order to prevent the use of prior experience in determining distance \cite{feldstein2020egocentric}. To assess immersion, we incorporated presence and embodiment questionnaires \cite{witmer2005factor} and modified it with a 5-point Likert scale ranging from "totally agree" to "totally disagree". It assesses participants' subjective experiences of body ownership in the virtual environment (e.g., whether they felt as if their own body was located where the virtual body was seen to be).

In our study, the captured real environments include two outdoor scenes (a park, a road) and two indoor scenes (a room, a corridor). For each environment, we captured a range of perspectives at varying eye heights from the lowest ($140 cm$) to the highest ($190 cm$) with $1 cm$ increment based on the average eye height ($165 cm$), revealed by previous research \cite{leyrer2011influence}. We then determine a lower eye height ($-25 cm$) and a higher eye height ($+25 cm$) based on the actual eye heights of each participant in the experiment. Fig. \ref{fig:setup} shows the setup used during the pilot study, and Fig. \ref{fig:environment} shows the captured real environments used in the experiment. Since stereoscopic vision does not significantly differ from monocular vision for perception beyond proximity, and to streamline the capture process for various height conditions, the captured environments were in monoscopic format. To prevent participants from using the movement of the target to guess the distance of the target, we prepared isolated clips of the real environments with the target positioned at fixed distances of $4 m$, $5 m$, and $6 m$ for each condition.

\textbf{Procedure.} 
After providing instructions to the participants, they were equipped with a HMD and underwent a brief calibration process. To familiarize themselves with the HMD, participants were given the opportunity to try it out before the start of the experiment. Each experiment consisted of a sequence of 36 distance estimation trials, during which participants were asked to view a pre-recorded virtual environment through the HMD and estimate the distance of a target object within the scene. The virtual environments consisted of four different settings (i.e, a park, a road, a room, and a corridor), and the target distances and virtual eye heights were varied across trials. To ensure that participants understood the task and to collect their distance estimates, the experimenter communicated with them verbally throughout the experiment. Participants did not receive feedback on the accuracy of their estimations during the experiment. Upon completion of the distance estimation sequence, participants removed the HMD and were instructed to complete the accompanying questionnaires and provide their consent.

\textbf{Results and findings.} 
We recruited 20 volunteers from the university, and the participants consisted of 14 males and 6 females. The sample was relatively homogeneous with an average age of 23.7 years old ($SD = 2.93$, 19-28 years old) and an average height of 168.9 cm ($SD = 9.72$, 154-183 cm). We set the level of significance to $\alpha$ = 0.05 and the power of the test to 1 - $\beta$ = 0.8.

No estimate given by the participants was removed from the analysis for being three standard deviations apart from the mean estimate. The analysis was conducted with a repeated-measures ANOVA with distance as the within-subjects factor, eye height and environment as between-subjects variables, and estimated distances as the dependent measure. Confirming our hypothesis, the influence of eye height on distance perception was significant (see Fig. \ref{fig:pilot}). Across the lower eye height ($M_{indoor}$ = .98, $SE_{indoor}$ = .035, $M_{outdoor}$ = .97, $SE_{outdoor}$ = .047), actual eye height ($M_{indoor}$ = .94, $SE_{indoor}$ = .033, $M_{outdoor}$ = .94, $SE_{outdoor}$ = .035), and higher eye height ($M_{indoor}$ = .82, $SE_{indoor}$ = .060, $M_{outdoor}$ = .79, $SE_{outdoor}$ = .050), estimated distances varied significantly ($p < .001$). With Fisher’s LSD tests, we found that the estimates given by the participants were different from the higher eye height significantly ($p < .001$), while a significant difference between the estimates under lower eye height and actual eye height was absent ($p = .25$). Furthermore, a similar effect was verified for body location ownership: a significant difference between the higher eye height and the other two conditions ($p < .001$), but no significant difference between the actual and the lower eye height ($p = .85$). The actual and lower height conditions showed very high responses of ownership, 4.3 and 4.2, on the 5-point Likert scales.

% After conducting an analysis using a repeated-measures analysis of variance (ANOVA) with distance ($4 m$, $5 m$, or $6 m$) as the within-subjects factor, eye height (lower, actual, or higher) and environment (indoor or outdoor) as between-subjects variables, and estimated distances as the dependent measure, we found that the influence of eye height on distance perception was significant ($p < .001$). We also observed a similar effect for body location ownership, with a significant difference between the higher eye height condition and the other two conditions, but no significant difference between the actual and lower eye height conditions. The actual and lower height conditions showed very high levels of body location ownership. Further details about the experimental design and analysis will be discussed in Section 4.2. 

% \subsection{Pilot Study}
% The goal of this pilot study is to investigate the hypothesis that eye height has a significant influence on perception and immersion in pre-captured real-world environments. Participants were asked to give a verbal estimation of the distance of a truncated cone-shaped target with a height of 10 cm. The dimensions of the object were not known to the participants to prevent the deduction of distance based on experience \cite{feldstein2020egocentric}.

% \section{Methodology}

\begin{figure*}[ht]
\centering
  \includegraphics[width=0.75\textwidth,keepaspectratio]{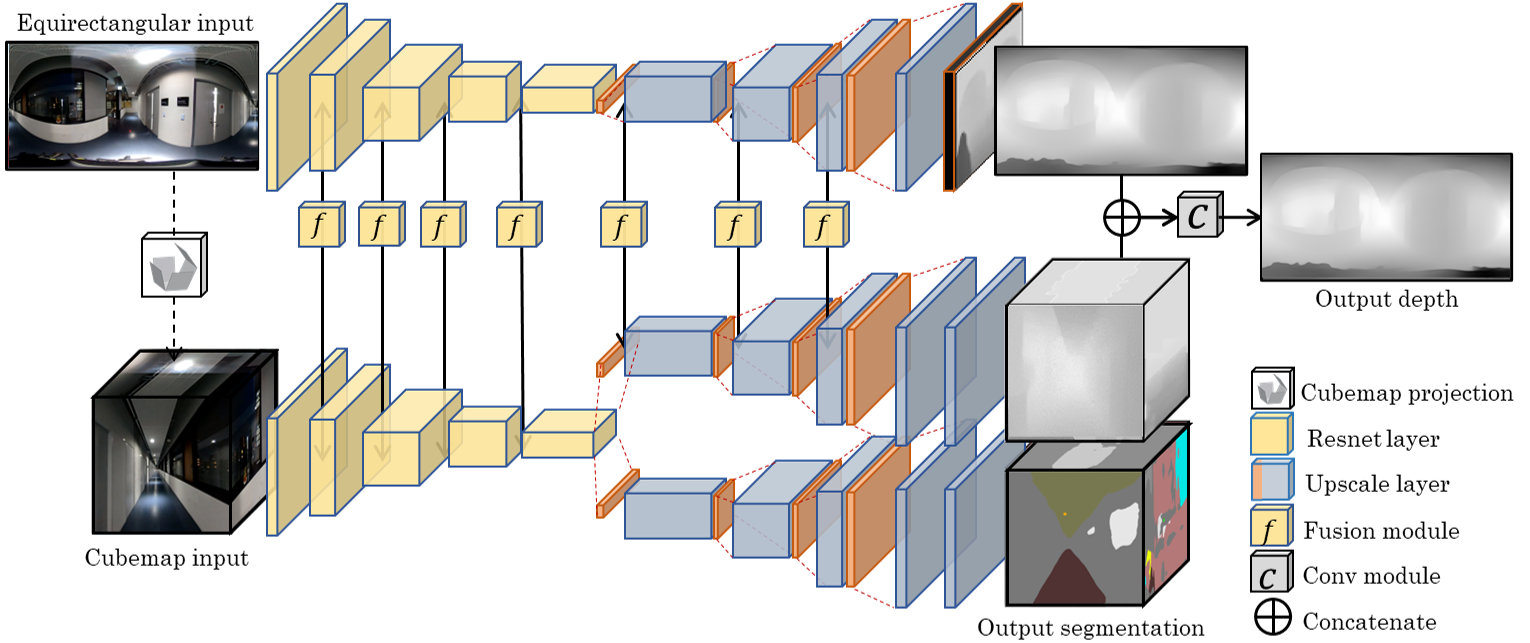}
    \vspace{-1em}
    \setlength{\belowcaptionskip}{-12pt}
  \caption{The overview of the proposed multitask network. %For 
  Depth estimation: the top and middle branches learn to predict the depth of the same scene with both equirectangular projection and cubemap projection, leading to sharper boundaries for local objects while maintaining consistent and smooth prediction for the entire scene with the extreme FOV. %For 
  Semantic segmentation: the middle and bottom branches with cubemap projection jointly learn to reveal the scene layout with a smaller FOV, providing %. This further provides 
  valuable guides and facilitating the %following 
  inpainting stage.}
  \label{fig:multitask}
\end{figure*}

% \vspace{-16pt} 

\section{Learning-based Eye Height Adaptation}

Motivated by the findings of the pilot study, we propose a novel learning-based approach for synthesizing novel views with varying eye heights from omnidirectional images. The pipeline is shown in Fig. \ref{fig:overview}). The system consists of two stages: depth estimation and image inpainting. After receiving the color input, the depth estimation stage uses a novel multitask network to simultaneously provides depth information and semantic segmentation guides from RGB input. In the inpainting stage, we improve existing LDI representation to enable high-quality inpainting results for omnidirectional images. By leveraging the semantic guides (see Fig. \ref{fig:ldi}), we generate visually convincing results at occluded regions. Finally, we merge the inpainted results back into the original LDI representation to render novel views with altered eye heights.

% for the subsequent image inpainting stage.we address the challenge of inpainting an equirectangular image with an extreme FOV and strong distortions into multiple local inpainting problems, allowing us to apply standard CNN models to generate visually convincing results. 
% In the depth estimation stage, we propose an architecture that employs both equirectangular and cubemap projections to regress the depth of the input image (see Fig. \ref{fig:multitask}). By simultaneously learning a semantic segmentation task in parallel with depth estimation, In the image inpainting stage, we 
% This allows the network to accurately estimate depth and generate precise semantic segmentation, facilitating the subsequent inpainting stage. By learning both tasks in parallel, our architecture is able to improve boundary estimation for local objects and provide useful semantic guides for inpainting. 

\subsection{Depth estimation and semantic segmenatation} 

We present a novel multitask architecture that simultaneously learns depth and semantic segmentation in two different formats (Fig. \ref{fig:multitask}). By simultaneously learning a semantic segmentation task in parallel with depth estimation, our model not only improves boundary estimation for local objects, but also facilitates the subsequent image inpainting stage by providing useful semantic guides. This approach offers a promising solution for synthesizing novel views from omnidirectional images.

\textbf{Multitask architecture.} To obtain accurate depth maps for omnidirectional images, we propose regressing dense global depth estimation from a single view equirectangular image in two different projections: equirectangular and cubemap. For the equirectangular input, the architecture has an encoder-decoder structure that progressively down-projects and up-projects to the original size. The advantage of directly learning depth estimation on the entire 360-degree input is that low spatial frequencies better represent global features such as structures and shapes in the scene. Coarse perception of the scene can be further exploited by a smoothness loss function to ensure that the learned depth is consistent and uniform. However, the disadvantage of projecting a sphere onto a flat 2D plane is the strong distortion introduced by uneven pixel densities. Distortion is stronger for sparse pixels near the poles and less prominent at the equator \cite{zioulis2019spherical}. To address this issue, we use rectangular filters with varying sizes at the first convolution layer along the vertical axis of the input equirectangular image. The encoder of this branch shares the same structure as ResNet-50, while the decoder consists of four up-projection blocks \cite{laina2016deeper}.

For the cubemap input, we first project the spherical image onto a cube to obtain cubemap faces. The use of cubemap projection in depth estimation is motivated by the desire to reduce distortion in the input image and provide higher spatial frequencies for improved shape and boundary detection of local objects. When directly using equirectangular images to learn depth estimation, details of local objects with steep gradient changes are usually omitted during the training process. By learning the depth estimation from both the equirectangular and cubemap projections, our model is able to learn complementary features from the same input. To encourage feature sharing and balance between the two branches, we use a fusion process and spherical padding \cite{wang2020bifuse} to connect the cube faces. This allows our model to adapt to weights trained for pinhole cameras and improve learning accuracy and efficiency. The final depth estimation is generated by projecting the output from the cubemap back onto the equirectangular projection and applying a convolution module.

We use $m_p$ and $m_f$ to represent feature maps from the equirectangular and cubemap branches, respectively. These maps are reprojected to $\hat{m_f}$ in equirectangular format and $\hat{m_p}$ in cubemap projection and fed into the next layer of the respective branches. In this layer, the reprojected maps are passed through a convolution layer ($C$) and added to the original feature maps. The result, $m_{p}+C(\hat{m_f})$, is then passed to the next layer of the equirectangular branch, while $m_{f}+C(\hat{m_p})$ is passed to the cubemap branch. This fusion process enables our model to learn complementary features from the two projections, improving the accuracy of the depth estimation.

To enhance the detection of depth discontinuities and improve the performance of the inpainting stage, our multitask architecture learns semantic segmentation from the cubemap input. We use the same encoder for both depth estimation and semantic segmentation to improve boundary recognition. Additionally, we train a separate decoder to generate semantic segmentation in the cubemap format. With a similar FOV of perspective images, the branch %allows as to 
utilizes abundant training data and pre-trained weights to improve the accuracy and efficiency of the entire training process of the proposed network.

% To make our model directly applicable to pre-captured real-world environments, we pre-train the semantic segmentation on the MIT ADE20K dataset \cite{zhou2017scene}, which includes perspective images. This pre-training helps to accelerate the training process for our proposed network.

\textbf{Loss Functions.}
In our model, we use supervised loss constraints for both the depth estimation and semantic segmentation tasks. For the depth estimation task, we use the inverse Huber loss function, which is defined in \cite{laina2016deeper}, as the optimizing objective $L_{Berhu}(d_i, \hat{d}_i)$:
\begin{equation}
L_{Berhu}(d_i, \hat{d}_i)=\begin{cases} |d_i-\hat{d}_i| & |d_i-\hat{d}_i| \geq c \\
                           \frac{(d_i-\hat{d}_i)^2+c^2}{2c} &  |d_i-\hat{d}_i|>c 
               \end{cases}
\end{equation}
where $d_i$ is the ground truth depth of the $i$th pixel, and $\hat{d}_i$ is the predicted depth of the $i$th pixel, and $c=\max(|d_i-\hat{d}_i|)/5$. The loss function $L_{Seg}$ for semantic segmentation is a cross-entropy loss between the estimated segmentation $\hat{s}_i$ and the result predicted with a pre-train network $S$. Combined, the total loss can be defined as:
\begin{equation}
L_{Total}=L_{Berhu}(d_i, \hat{d}_i)+L_{Seg}(s_i, \hat{s}_i)
\end{equation}

\subsection{Context-aware inpainting}
In the inpainting stage, we improve upon the existing LDI approach \cite{shih20203d} by incorporating guidance from the omnidirectional-aware depth and semantic segmentation information. Our method can therefore synthesize realistic textures for occluded regions, leading to more natural and realistic images. This is especially beneficial for pre-captured environments, where the ability to adapt to different eye heights is crucial for creating a convincing experience.

\begin{figure}[htb]
\centering
  \includegraphics[width=0.47\textwidth,keepaspectratio]{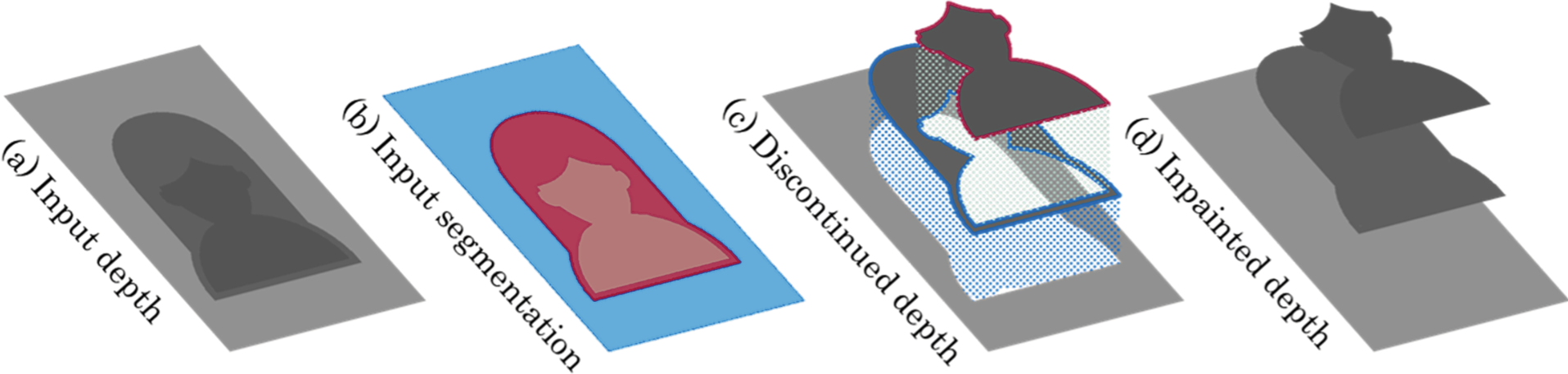}
  \vspace{-1em}
  \setlength{\belowcaptionskip}{-12pt}
  \caption{Illustration of the proposed process of LDI. For fully connected input depth image (a), we first separate the input according to segmentation information (b), resulting in foreground (red) and background regions (gray). After missing pixels are inpainted and no holes are left for each layer, results are then merged into the final LDI.}
  \label{fig:ldi}
\end{figure}

\textbf{Backbone.} LDIs are effective ways to synthesize novel viewpoints from color and depth information. LDIs can handle arbitrary numbers of layers, which allows them to represent complex scenes. Each LDI pixel contains color information, a corresponding depth value, and pointers to horizontal and vertical neighboring pixels. In the case of depth discontinuities, LDI pixels will have zero neighboring pixels in the relevant cardinal direction.

To use an LDI (Fig. \ref{fig:ldi}), we first initialize it with a single layer that is fully connected in all directions. We then utilize the estimated semantic segmentation information and the previously-generated depth map to identify depth discontinuities and group them into simple and connected local edges. Subsequently, we disconnect LDI pixels across these edges and apply inpainting networks to fill in the occluded regions for both the depth map and color image. After %the completion of 
inpainting, we merge the synthesized pixels back into the LDI until all the local edges have been resolved. %Figure \ref{fig:ldi} illustrates this process.

\textbf{Multi-layer inpainting.} Our goal is to use LDIs to inpaint occluded regions in an omnidirectional image and render views from a different eye height. This will allow us to synthesize novel views that closely resemble the real environment from the specified viewpoint.

To identify the regions that require inpainting, it is essential to accurately identify depth discontinuities in the input image. Traditional methods employ thresholding algorithms, which result in blurred boundaries across multiple pixels \cite{shih20203d}. In addition, generating photo-realistic output from existing LDIs requires precise pairing of depth and color images, which can be challenging. Although CNNs have been utilized to address this issue for single-image inputs, models designed for images captured with pinhole cameras often produce suboptimal results when synthesizing depth information, resulting in inconsistent novel views with artifacts \cite{attal2020matryodshka}. To overcome these challenges and improve the accuracy of the generated LDIs, we employ the proposed omnidirectional-aware multitask network that generates guiding semantic segmentation and detects discontinuities in the estimated depth maps (Fig. \ref{fig:ldi} (c)). We create a binary mask, labeling depth discontinuities as 1 and the remaining pixels as 0. We then merge adjacent discontinuities into linked local edges and use connected component analysis to prevent merging across discontinuities. Finally, we exclude local edges with less than ten pixels in length to obtain the regions for the inpainting process.

To perform inpainting of both depth and color information, we utilize a standard encoder-decoder architecture with U-Net and partial convolution for the depth inpainting network, as proposed in \cite{liu2018image}. The color inpainting network is structured similarly, with the same number of layers. The depth inpainting network takes the contextual depth and local edges as input, while the color inpainting network takes the contextual color image and local edges as input. The training objectives for each network are as follows:
\begin{equation}
\vspace{-1em}
L_{Depth}=\frac{1}{N} ||S \odot (d_i, \Tilde{d}_i) || 
\end{equation}
% \vspace{-1em}
\begin{equation}
L_{Color} = \alpha (\frac{1}{N} ||S \odot (c_i, \Tilde{c}_i) ||) + \beta L_{Perceptual}
\end{equation}
where $d_i$ is the ground truth depth of the $i$th pixel, $\Tilde{d}_i$ is the inpainted depth of the $i$th pixel, $c_i$ is the ground truth color of the $i$th pixel, and $\Tilde{c}i$ is the inpainted color of the $i$th pixel. $S$ is a binary mask that describes the contextual region, $\odot$ denotes the Hadamard product, and $L_{Perceptual}$ is the loss function for the color inpainting task. It is obtained using the output of layers from a pre-trained VGG-16 model 
% \cite{simonyan2014very}
. The color inpainting network is trained on COCO-2017 \cite{lin2014microsoft}, while the depth inpainting network is trained on MegaDepth \cite{li2018megadepth}.

\textbf{Eye height adaptation.}
To modify the viewpoint of a pre-captured environment with a different eye height, we use a vertical geometric model to reproject the LDIs to the desired viewpoint, as depicted in Fig. \ref{fig:teaser}. Specifically, we apply the vertical spherical model introduced in \cite{zioulis2019spherical} to transform the original view $j$ to the altered view $k$ with a vertical disparity $d$. The transformation is done in polar coordinates, where each point $p$ at $(x,y,z)$ in Cartesian coordinate is represented by its longitude $\phi$ and latitude $\theta$. The radial distance $r$ (i.e., depth value) of a point is given by $\sqrt{x^2 + y^2 + z^2}$, and the vertical distance is defined as $\delta = (\phi_j - \phi_k, \theta_j - \theta_k)$. As we only need to adapt the eye height, we only consider vertical disparity $d = (0, dy, 0)$, and the disparity is reduced to $\delta = (\frac{\partial\phi}{\partial y}, \frac{\partial\theta}{\partial y})$.

To render a target view $\hat{k}$ from the source input $j$, each pixel $p=(\phi,\theta)$ on the equirectangular image is a function of the vertical disparity $d$ and the radial distance $r$. Since we already have the depth and color information from LDIs, we can compute the target frame $\hat{k}$ with a function: 
\begin{equation}
\hat{k}(p)= \Gamma_{j \rightarrow \hat{k}}(\Tilde{d},d_{j \rightarrow k}, j(p))
\end{equation}

%%%%%%%%%%%%%%%%%%%%%%%%%%%%%%%%%%%%%%%%%%%%%%%%%%%%%%%%%%%%%%%%%
%%%%%%%%%%%%%%%%%%%%% START OF EXPERIMENTS %%%%%%%%%%%%%%%%%%%%%%
%%%%%%%%%%%%%%%%%%%%%%%%%%%%%%%%%%%%%%%%%%%%%%%%%%%%%%%%%%%%%%%%%

\section{Experimental Results}
\subsection{Implementation Details}
We have implemented our proposed multitask network using the PyTorch framework 
% \cite{paszke2017automatic}
, and trained it on a single Nvidia RTX 2080Ti graphics card, using data from the Depth360 dataset \cite{feng2022360}. During training, we employed the Adam optimizer 
% \cite{kingma2014adam} 
with a learning rate of 3e-4, and used a batch size of 1 due to graphics memory constraints. Our equirectangular branch was initialized with Xavier initialization \cite{glorot2010understanding}, while the cubemap branch was initialized with ImageNet pretrained weights. Our approach takes around $150ms$ to predict depth maps and semantic segmentation for a single equirectangular image. For the contextual inpainting networks, we trained the depth inpainting network on MegaDepth \cite{li2018megadepth} for 5 epochs, while the color inpainting network was trained on the MS-COCO \cite{lin2014microsoft} for 10 epochs. We used $\alpha = 1$ and $\beta = 0.05$ as the parameters for the inpainting process.

\begin{figure}[t]
\centering
  \includegraphics[width=0.45\textwidth,keepaspectratio]{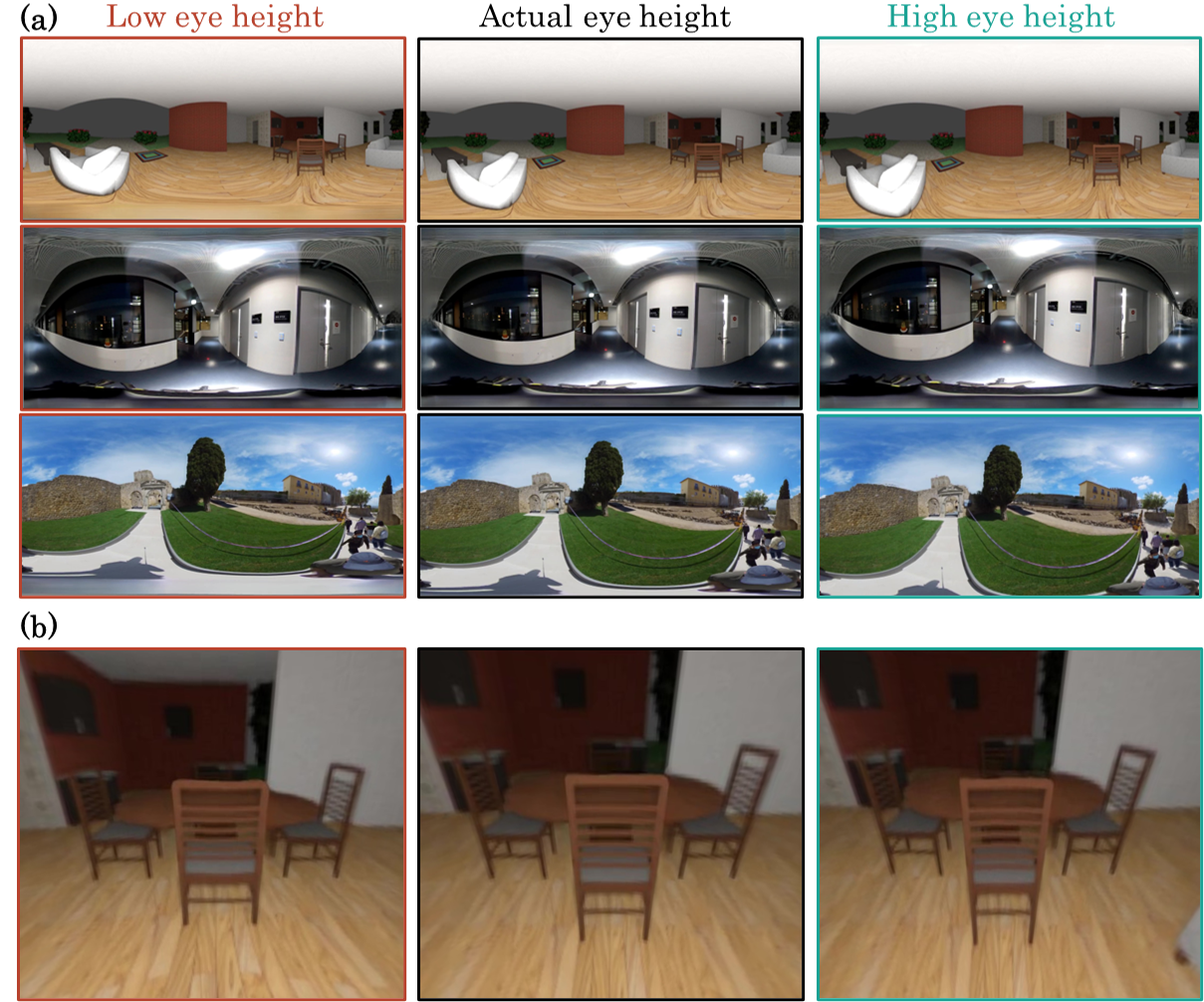}
  \vspace{-1em}
  \setlength{\belowcaptionskip}{-12pt}
  \caption{Results of the proposed eye height adaptation approach. (a) shows the omnidirectional visuals when adapted to a higher eye height ($+ 25 cm$) and a lower eye height ($- 25 cm$). (b) shows perspective visuals when viewed with a smaller FOV (i.e., HMD).}
  \label{fig:qualitative1}
\end{figure}

\begin{figure}[htb]
\centering
  \includegraphics[width=0.45\textwidth,keepaspectratio]{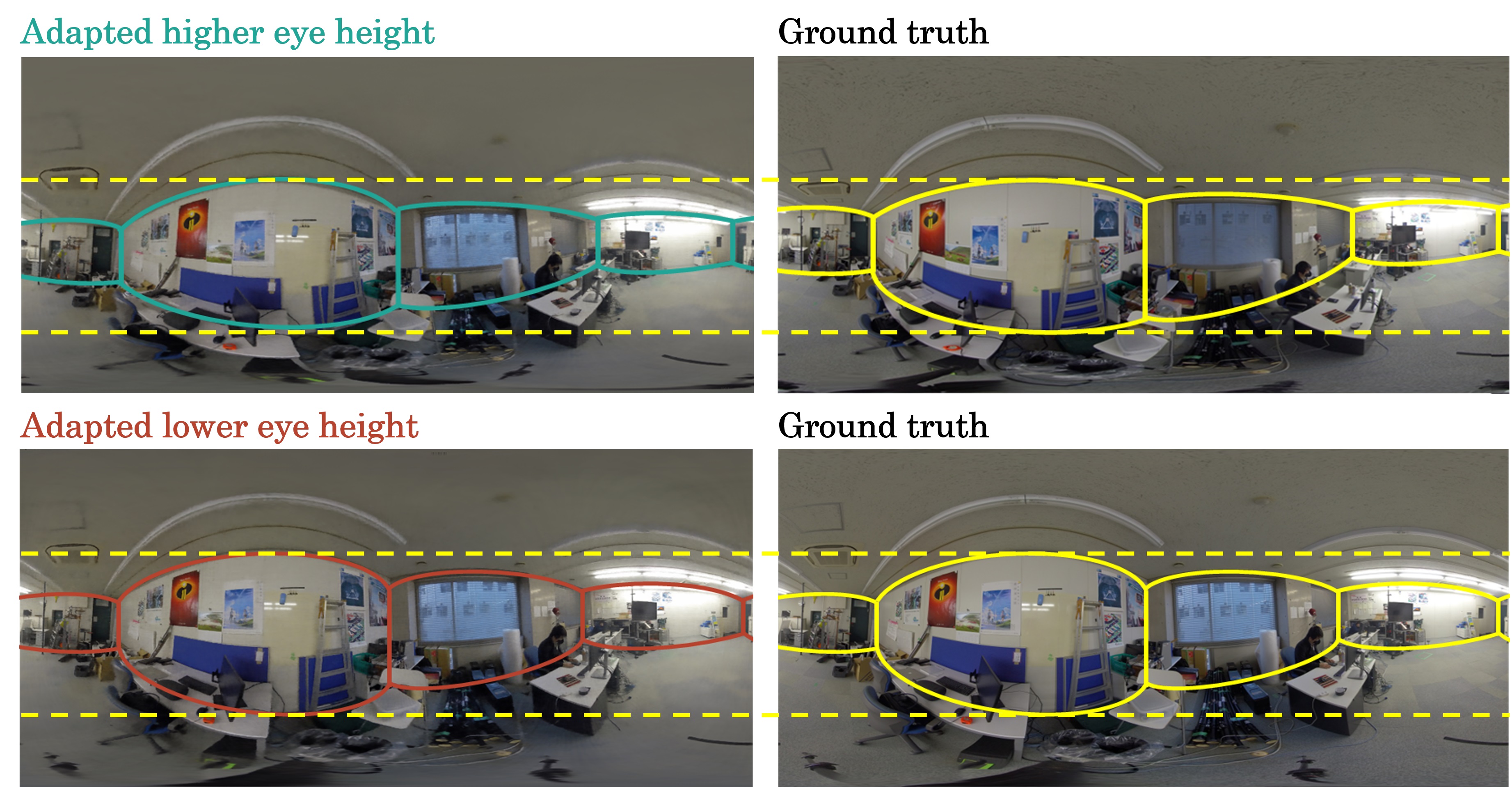}
  \vspace{-1em}
  \setlength{\belowcaptionskip}{-9pt}
  \caption{Comparisons of scene structures between adapted views and captured ground truth showed that the generated visuals at both higher ($+25 cm$) and lower ($-25 cm$) adapted eye heights matched the ground truth scene structures.}
  \label{fig:structure}
\end{figure}

\begin{figure}[htb]
\centering
  \includegraphics[width=0.45\textwidth,keepaspectratio]{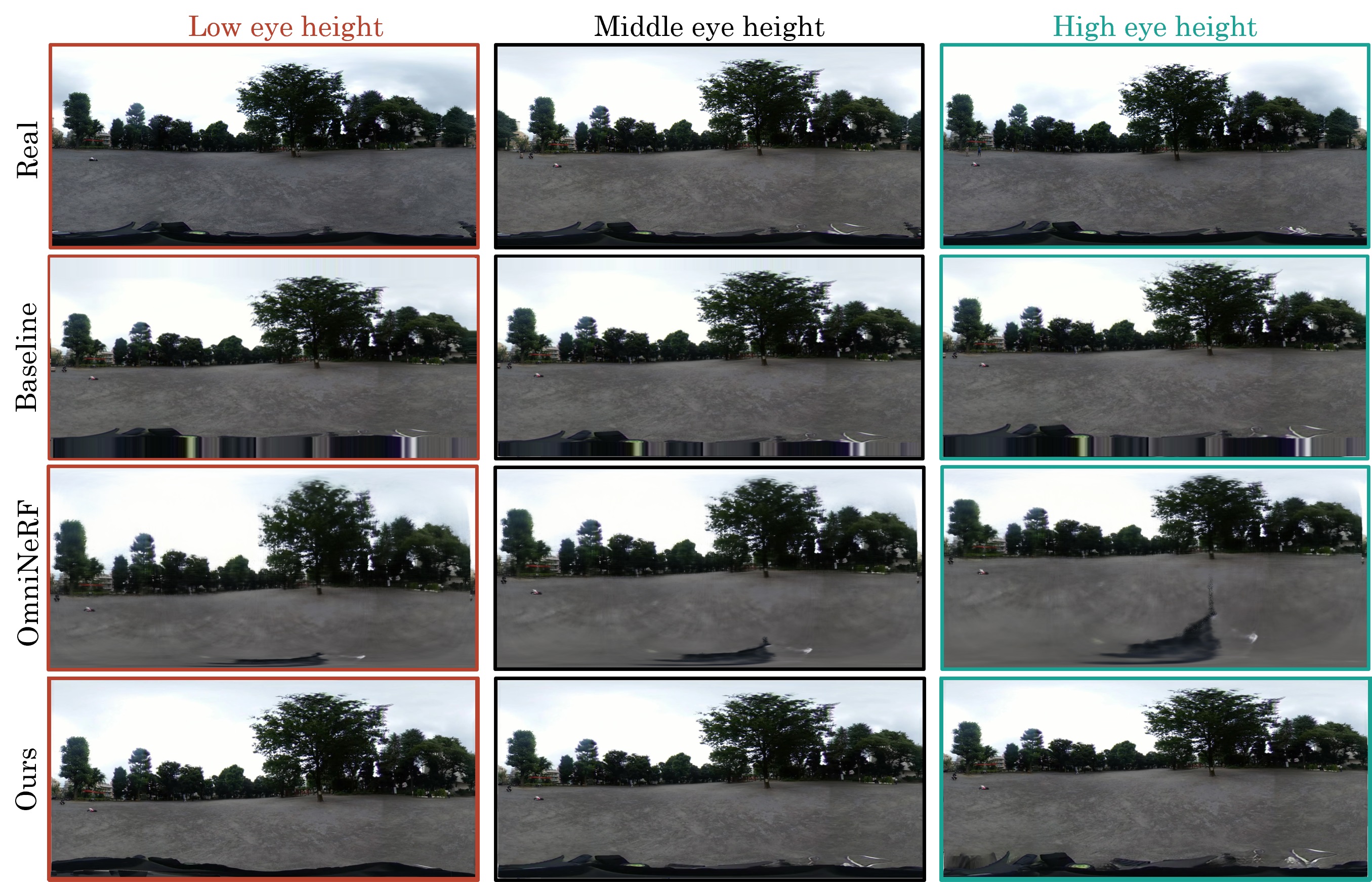}
  \vspace{-1em}
  \setlength{\belowcaptionskip}{-12pt}
  \caption{Qualitative comparisons against a baseline method of parallax mapping that completes the occluded regions with bilinear interpolation, and a state-of-the-art NeRF-based method designed for novel view synthesis from omnidirectional input.}
  \label{fig:qualitative2}
\end{figure}

\begin{figure}[htb]
\centering
  \includegraphics[width=0.45\textwidth,keepaspectratio]{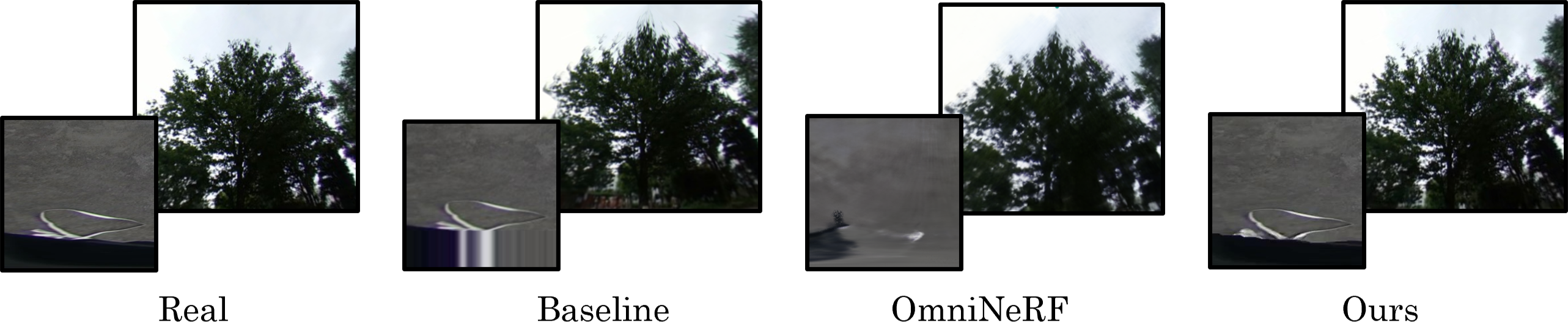}
  \vspace{-1em}
  \setlength{\belowcaptionskip}{-12pt}
  \caption{Close-up views for the output with adapted eye heights. Compared to other approaches, the proposed method generates more natural and clearer visuals for both local regions and image boundaries with strong distortion.}
  \label{fig:qualitative2.5}
\end{figure}

\begin{figure}[htb]
\centering
  \includegraphics[width=0.42\textwidth,keepaspectratio]{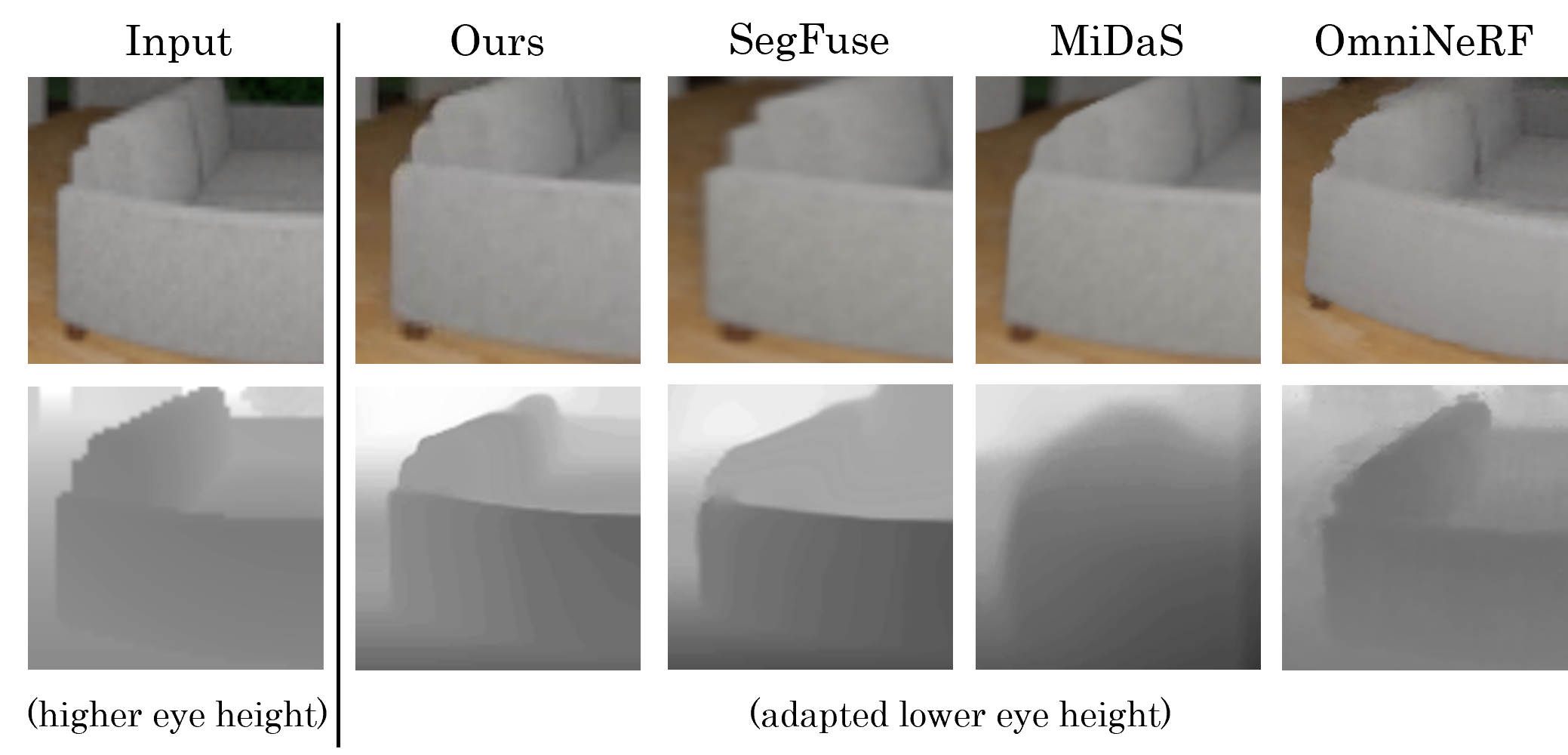}
  \vspace{-1em}
  \setlength{\belowcaptionskip}{-12pt}
  \caption{Comparisons between different depth estimation approaches. For the same input, we swap the proposed multitask network to different depth estimation models for eye height adaptation.% and compare the output. 
  When decreasing the eye height by $25cm$, the proposed method shows the sharpest edges for local objects and the most consistent depth estimation.}
  \label{fig:qualitative4}
\end{figure}

\begin{table}[htb]
\centering

\caption{\textbf{Quantitative comparison} on the SynDepth360 dataset \cite{feng2022360}.}
\resizebox{0.63\columnwidth}{!}{%
\begin{tabular}{c*{3}{>{$}c<{$}}}
 \toprule
  Method & SSIM \uparrow & PSNR \uparrow & LPIPS \downarrow\\
 \midrule
    Baseline & 0.8347 & 23.74 & 0.0982 \\
    OmniNeRF \cite{hsu2021moving} & 0.8528 & 26.74 & 0.0824 \\
    Ours & \textbf{0.8776} & \textbf{27.19} & \textbf{0.0736} \\
 \bottomrule
 \label{table: qualitative1}
 % \vspace{-1em}
\end{tabular}}

%\vspace{0.2cm}

\caption{\textbf{Average computation time and memory requirement per image.} Every approach is evaluated using Nvidia RTX 2080Ti and Intel i7-7800X, using input with a resolution of 1024 $\times$ 512.}
\resizebox{0.75\columnwidth}{!}{%
\begin{tabular}{c*{2}{>{$}c<{$}}}
 \toprule
  Method & Computation\; time\; [s] & Memory\; cost \;[MB]\\
 \midrule
OmniNeRF \cite{hsu2021moving} & > 250,000 & 13.6\\
Ours & \textbf{0.450} & \textbf{1.42}\\ 
 \bottomrule
 \label{table: qualitative2}
\end{tabular}}

%\vspace{0.2cm}

\caption{\textbf{Quantitative results} of depth estimation on Matterport3D dataset \cite{chang2017matterport3d}, SunCG dataset \cite{wang2018self}, and Depth360 dataset \cite{feng2022360}.}
\resizebox{0.95\columnwidth}{!}{%
\begin{tabular}{c*{5}{>{$}c<{$}}}
\toprule
  Method  & \text{RMSE} \downarrow & \text{RMSE log} \downarrow & \text{$\delta < 1.25$} \uparrow & \text{$\delta < 1.25^2$} \uparrow & \text{$\delta < 1.25^3$} \uparrow \\
     \midrule
\multicolumn{6}{c}{The Matterport3D dataset \cite{chang2017matterport3d} (Real domain, indoor settings.)} \\
     \midrule
 BiFuse \cite{wang2020bifuse} & 0.6259 & 0.1134 & 84.52\% & 93.19\% & 96.32\%\\ % OK
 SegFuse \cite{feng2022360} & 0.6029 & 0.1100 & 84.60\% & 93.14\% & 96.28\%\\ % OK
 MiDaS \cite{ranftl2019towards} & 0.7641 & 0.1420 & 77.05\% & 88.94\% & 95.77\%\\ % OK
 MegaDepth \cite{li2018megadepth} & 0.7845 & 0.1502 & 69.50\% & 87.94\% & 94.21\%\\ % OK 
 Ours & \textbf{0.5880} & \textbf{0.0986} & \textbf{85.17}\% & \textbf{93.45\%} & \textbf{96.85\%}\\ % OK
     \midrule
\multicolumn{6}{c}{The SunCG dataset \cite{wang2018self} (Synthetic domain, indoor settings.)} \\
     \midrule
 BiFuse \cite{wang2020bifuse} & 0.2596 & 0.0443 & \textbf{95.90\%} & 98.23\% & 99.07\%\\ % OK
 SegFuse \cite{feng2022360} & 0.2540 & 0.0427 & 95.32\% & 98.34\% & 99.10\%\\ % OK
 MiDaS \cite{ranftl2019towards} & 0.3244 & 0.0730 & 89.90\% & 96.62\% & 95.44\%\\ % OK
 MegaDepth \cite{li2018megadepth} & 0.4041 & 0.0845 & 84.06\% & 93.92\% & 94.85\%\\ % OK 
 Ours & \textbf{0.2490} & \textbf{0.0425} & 95.57\% & \textbf{98.45\%} & \textbf{99.26\%}\\ % OK
      \midrule 
\multicolumn{6}{c}{The Depth360 dataset \cite{feng2022360} (Real domain, outdoor settings.)} \\
     \midrule
 BiFuse \cite{wang2020bifuse} & 5.0725 & 0.8316 & 40.13\% & 59.17\% & 67.92\%\\ % OK
 SegFuse \cite{feng2022360} & \textbf{4.0442} & \textbf{0.7777} & \textbf{82.26\%} & \textbf{91.35\%} & \textbf{94.22\%}\\ % OK
 MiDaS \cite{ranftl2019towards} & 6.0132 & 0.9574 & 52.26\% & 58.73\% & 65.31\%\\ % OK
 MegaDepth \cite{li2018megadepth} & 6.7320 & 0.9863 & 48.33\% & 61.75\% & 64.82\%\\ % OK
 Ours & 4.0544 & 0.7965 & 81.07\% & 90.89\% & 94.18\%\\ % OK
 \bottomrule
 \label{table: qualitative3}
\end{tabular}}
\vspace{-18pt}%
\end{table}
  % \vspace{-1em}
% \vspace{-12pt} 

\subsection{Qualitative Evaluation}
We qualitatively evaluates 
%This section presents the qualitative results of 
our proposed eye height adaptation method for pre-captured immersive content. Fig. \ref{fig:qualitative1} displays the visual results obtained when our method was tested on unseen equirectangular images with both indoor and outdoor settings. As shown in Fig. \ref{fig:qualitative1} (b), our method successfully generates new perspectives with varying eye heights. To showcase the effectiveness and accuracy of our method, we captured the ground-truth at respective eye heights using a 360-degree camera. As demonstrated in Fig. \ref{fig:structure}, the adapted views from different eye heights match the ground-truth with high accuracy, demonstrating the efficacy of our approach.

In addition, we conducted a comparative analysis of our method against a baseline approach and a state-of-the-art NeRF-based method called OmniNeRF \cite{hsu2021moving}. The baseline approach utilizes parallax mapping based on the new viewpoint and the depth estimated by our multitask network, and then employs bilinear interpolation for occluded pixels to complete the view. The results of the experiment are presented in Fig. \ref{fig:qualitative2} and Fig. \ref{fig:qualitative2.5}. Our proposed method generates visually compelling results with sharper details without the need for re-training for each scene over prolonged periods. On the other hand, although NeRF is capable of synthesizing images at arbitrary resolution through its implicit formation, the experiment showed low visual fidelity due to sparse sampling in a single omnidirectional image. Nonetheless, it is worth noting that NeRF methods excel in rendering specular surfaces and reflections with its ray tracing capability, while image-based methods encounter challenges in achieving comparable realism in this aspect. 

\subsection{Quantitative Evaluation}
In this section, we assess the accuracy of our proposed approach in adapting input RGB images to different eye heights for omnidirectional images and compare it with state-of-the-art methods. To mitigate potential lighting and scene configuration variations when capturing the same real-world scene with different eye heights, we utilized SynDepth360 \cite{feng2022360}, a small-scale synthetic omnidirectional dataset with 3D models, to generate novel perspectives with virtual camera placements altered for evaluating different methods. We employed SSIM, PSNR, and LPIPS metrics \cite{shih20203d} to quantify output quality % the precision of the output compared to the ground truth 
as shown in %. The results are presented in 
Table \ref{table: qualitative1}.

\textbf{Run-time efficiency.} We evaluate the processing time and memory requirements from inputting the image to render adapted views in Table \ref{table: qualitative2}. For our method, it means passing the input through the feed-forward models (Fig. \ref{fig:overview}); for NeRF, it means the system augments the single input and learn volumetric representation/rendering. In addition to greatly improved run-time efficiency, our LDI-based method is memory efficient, making it a viable option for future applications on commercial HMDs with independent processors.

% \subsection{Ablation Study}
\textbf{Network evaluation.}
We aim to evaluate the performance of our proposed multitask network for accurate depth estimation, which is crucial for generating the LDI required for rendering adapted views. To assess our approach, we adopt commonly used depth prediction metrics from literature to evaluate robustness and accuracy against state-of-the-art techniques. Table \ref{table: qualitative3} presents the results of our proposed method alongside those of SegFuse \cite{feng2022360} and BiFuse \cite{wang2020bifuse}, which are omnidirectional-aware methods, as well as perspective-based methods MiDaS (v2.1) \cite{ranftl2019towards} and MegaDepth \cite{li2018megadepth}. All models were trained using their released code bases for 20 epochs. We further demonstrate the results for adapted eye height with swapped depth estimation components in Fig. \ref{fig:qualitative4}. While our multitask network outperforms all other methods in predicting the layout of indoor omnidirectional images, the depth estimation learned through cubemap projection can suffer from boundary issues in outdoor settings, as previously pointed out in \cite{feng2022360}. Despite this limitation, our method still shows comparable accuracy to SegFuse in outdoor settings. Overall, our method achieves high accuracy in depth estimation and effectively achieves natural and accurate eye height adaptation.

%%%%%%%%%%%%%%%%%%%%%%%%%%%%%%%%%%%%%%%%%%%%%%%%%%%%%%%%%%%%%%%%%
%%%%%%%%%%%%%%%%%%%%%% START OF USER STUDY %%%%%%%%%%%%%%%%%%%%%%
%%%%%%%%%%%%%%%%%%%%%%%%%%%%%%%%%%%%%%%%%%%%%%%%%%%%%%%%%%%%%%%%%

\subsection{User Study}
The goal of this %user 
study was to examine the effectiveness of the proposed eye height adaptation approach for pre-captured immersive environments. %In these experiments, we 
These experiments explore the influence of manipulated eye heights on perception and immersion %, specifically, 
using verbal estimations of egocentric distances and body location ownership questionnaires.

\textbf{Participants and apparatus.}
In this study, we recruited a sample of 22 participants from the university, including 15 males and 7 females. The average age of the participants was 23.6 years old ($SD = 3.1$), and the average height was 166.9 cm ($SD = 9.17$). All participants had normal visual acuity and were comfortable using HMDs. The experiment was implemented in Unity3D and played back wirelessly on an Meta Quest 2 HMD using Air Link. The system used an Nvidia RTX 2080 Ti GPU, an Intel i7-7800X CPU, and 32GB RAM. The HMD had a refresh rate of 90 HZ and a field of view of 90$^{\circ}$ $\pm$ 5$^{\circ}$ horizontally and 93$^{\circ}$ $\pm$ 5$^{\circ}$ vertically, based on individual participant fitting. We used the same questionnaires as in the pilot study to assess perceived immersion and the same metric to compare the discrepancy between estimated and actual distance.

\textbf{Design and procedure.}
% Our primary study employed a within-subjects design, with manipulated eye height as the independent variable. In contrast to the pilot study, we utilized the HMD's built-in tracking capability to obtain the participant's actual eye height during the experiment. As a result, we rendered three conditions of the same scene based on the user's actual eye height: a normal eye height corresponding to the participant's actual eye height, a low eye height 25 cm lower than the calculated actual eye height, and a high eye height 25 cm higher than the calculated actual eye height. To minimize the impact on distance estimation accuracy and showcase the proposed method's effectiveness, we provided four environments with similar conditions to the pilot study, two indoor and two outdoor scenarios.
In our main study, we used a within-subjects design to manipulate eye height as the independent variable. Unlike the pilot study, we used the HMD's tracking capability to obtain the participant's actual eye height. We used the lowest capture (140 cm) as the input to the pipeline, and rendered monoscopic views from 141 cm to 190 cm. During the study, we prepare adapted, low ($-25 cm$), and high ($+25 cm$) eye heights for each user. We tested this method's effectiveness by providing four similar environments to the pilot study, two indoor and two outdoor scenarios, while minimizing the impact on distance estimation accuracy.

During the experiment, the experimenter first provided instructions to each participant and answered any questions until the participant had a clear understanding of the task. Before equipping the HMD, the experimenter showed the participant a one-meter ruler to ensure their %they had a good 
understanding of distance. After loading each condition, participants had time to freely explore the environment until they were ready to verbally estimate the distance to the target. This was crucial for the post-experiment presence questionnaire, which assessed subjective body ownership and immersion. On average, each participant completed the experiment in approximately 20 minutes.

\begin{figure}[htb]
\centering
  \includegraphics[width=0.45\textwidth,keepaspectratio]{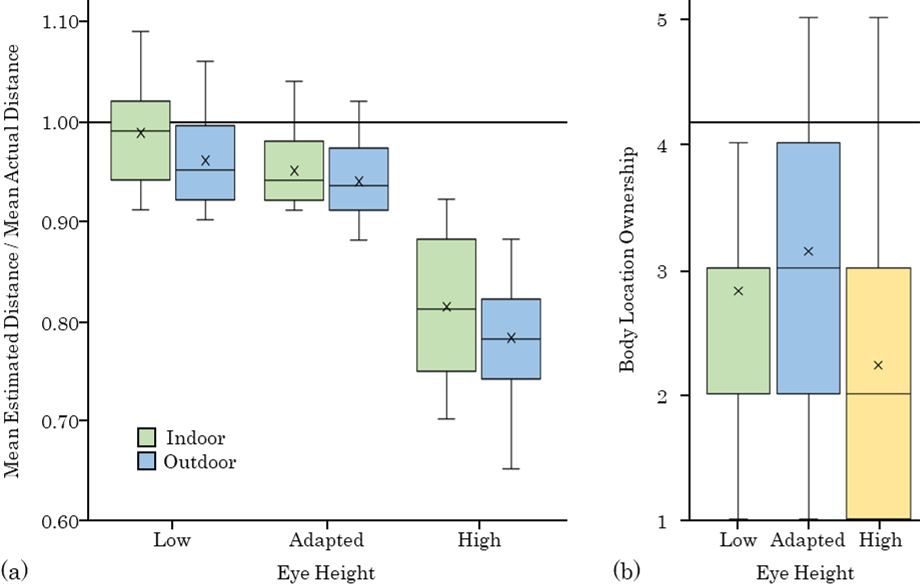}
  \vspace{-1em}
  \caption{User study results in distance perception and immersion.}
  \label{fig:study}
\end{figure}

\textbf{Results and general discussion.}
After conducting an analysis using a repeated-measures ANOVA with manipulated eye height (low, adapted, or high) and environment (indoor or outdoor) as the between-subjects factor, and estimated distances as the dependent measure, we confirmed the similar result to the pilot study that the adapted eye height significantly influenced distance perception (refer to Fig. \ref{fig:study}). The estimated distances varied significantly for low eye height ($M_{indoor}$ = .99, $SE_{indoor}$ = .049, $M_{outdoor}$ = .96, $SE_{outdoor}$ = .046), adapted eye height ($M_{indoor}$ = .95, $SE_{indoor}$ = .039, $M_{outdoor}$ = .94, $SE_{outdoor}$ = .039), and high eye height ($M_{indoor}$ = .81, $SE_{indoor}$ = .066, $M_{outdoor}$ = .78, $SE_{outdoor}$ = .055), with the effect being statistically significant ($p < .001$). The outdoor conditions show significant ($p < .001$) distance underestimation when compared to indoor conditions. This aligns with the findings from previous research \cite{masnadi2022effects}. Regarding body location ownership, our findings show that the adapted eye height conditions resulted in improved immersion responses with an average of 3.14 compared to low eye height ($M = 2.81$) and high eye height ($M = 2.22$). Although a significant influence was still observed when the manipulated eye height was higher than the user's actual eye height ($p < .001$), no significant effect on immersion was confirmed when the manipulated eye height was lower than the actual eye height of the user, consistent with both the pilot study and previous research. This finding can be explained by the abundance of daily actions people perform to lower their eye height, while actions to increase eye height are less common. In summary, our approach effectively improves distance perception and provides better immersion with adapted eye height when the environment is captured with a higher declination angle.

  \vspace{-1em}
\vspace{-8pt} 

\section{Conclusion and Future Work}
In this paper, we present a learning-based eye height adaptation method that generates corrected views from pre-captured immersive environments based on the user's actual eye height during playback. We first conducted a pilot study to confirm the hypothesis that different eye heights significantly influence distance perception and immersion. Subsequently, we propose a learning-based approach with a novel multitask architecture that learns to predict depth and semantic segmentation for omnidirectional images with high accuracy. By utilizing layered depth image representation and image inpainting to generate views with altered eye heights, our approach efficiently synthesizes natural-looking visuals. Evaluation against state-of-the-art approaches demonstrates the effectiveness of our proposed method, while a user study confirms the improvements in perception and immersion. 
% We believe that our approach can be directly applied to existing pre-captured immersive contents, enhancing the user experience. 
Future work will explore networks with better capability to generate high-resolution results for mixed reality, as well as the incorporation of efficient NeRF algorithms to improve accuracy in environments with specular surfaces. Finally, while this method can be applied to highly dynamic scenes on a per-frame basis, exploiting temporal and geometric information in videos is a another promising direction to ensure visual consistency.

\textbf{Acknowledgements}
This research is supported in part by JSPS KAKENHI (ref: JP21H05054) and the EPSRC NortHFutures project (ref: EP/X031012/1).

% The authors wish to thank A, B, and C. This work was supported in part by
% a grant from XYZ.}

%\bibliographystyle{abbrv}
\bibliographystyle{abbrv-doi}

\bibliography{template}
\end{document}